\documentclass[
	a4paper, 
	10pt, 
	unnumberedsections, 
	twoside, 
]{LTJournalArticle}

\runninghead{} 

\footertext{\textit{arXiv preprint}} 

\title{TWICE Dataset: Digital Twin of Test Scenarios in a  Controlled Environment} 

\author{%
	Leonardo Novicki Neto\textsuperscript{1,2}\thanks{Corresponding author: \href{mailto:novicki@ufpr.br}{novicki@ufpr.br}\\   October 5, 2023}, Fabio Reway\textsuperscript{1}, Yuri Poledna\textsuperscript{1}, Maikol Funk Drechsler\textsuperscript{1},\\ Eduardo Parente Ribeiro \textsuperscript{2}, Werner Huber \textsuperscript{1} and Christian Icking \textsuperscript{3}
}

\date{\footnotesize\textsuperscript{\textbf{1}}CARISSMA Institute of Automated Driving (C-IAD)\\ \textsuperscript{\textbf{2}}Federal University of Paraná\\ \textsuperscript{\textbf{3}}FernUniversität in Hagen}

\renewcommand{\maketitlehookd}{%
	\begin{abstract}
		\noindent 
Ensuring the safe and reliable operation of autonomous vehicles under adverse weather remains a significant challenge. 
To address this, we have developed a comprehensive dataset composed of sensor data acquired in a real test track and reproduced in the laboratory for the same test scenarios.
The provided dataset includes camera, radar, LiDAR, inertial measurement unit (IMU), and GPS data recorded under adverse weather conditions (rainy, night-time, and snowy conditions). 
We recorded test scenarios using objects of interest such as car, cyclist, truck and pedestrian -- some of which are inspired by EURONCAP (European New Car Assessment Programme). 
The sensor data generated in the laboratory is acquired by the execution of simulation-based tests in hardware-in-the-loop environment with the digital twin of each real test scenario. 
The dataset contains more than 2 hours of recording, which totals more than 280GB of data.
Therefore, it is a valuable resource for researchers in the field of autonomous vehicles to test and improve their algorithms in adverse weather conditions, as well as explore the simulation-to-reality gap. 
The dataset is available for download at: \texttt{https://twicedataset.github.io/site/}
	\end{abstract}
}

\begin{document}

\maketitle 

\section{Introduction}
\begin{figure*}[ht!]
	\centering
	\includegraphics[width=\textwidth]{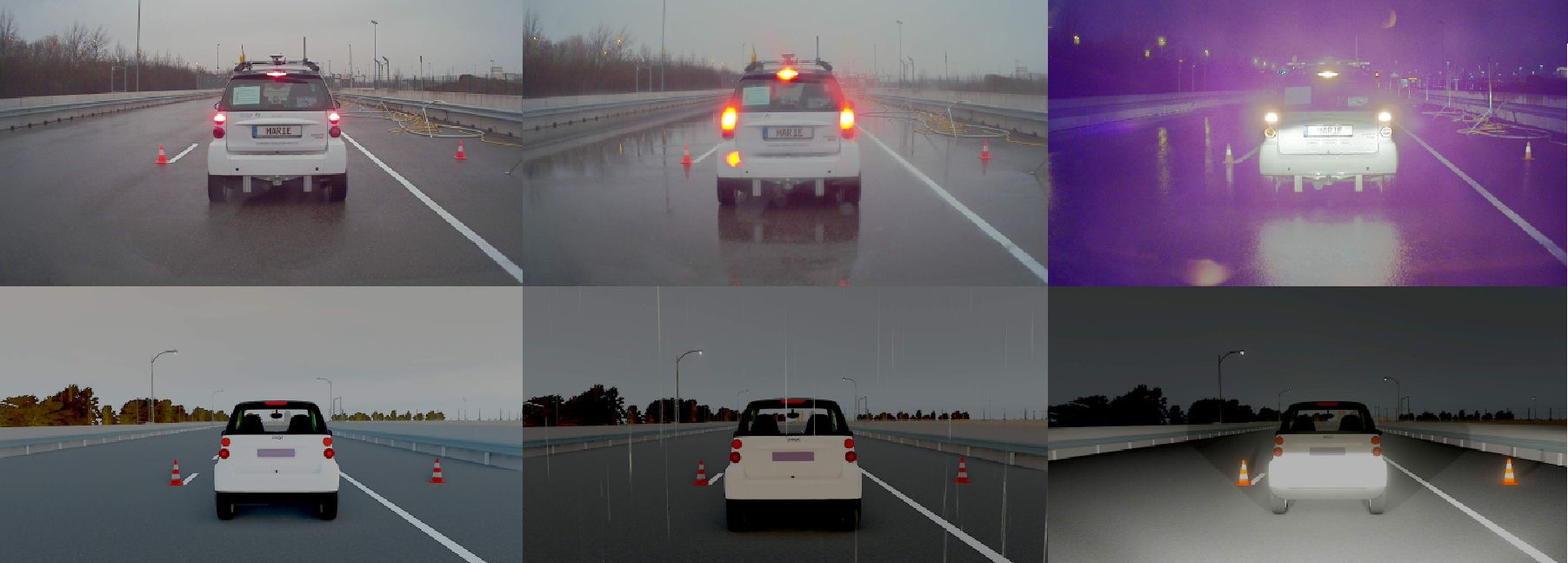}
	\caption{Real (top) and virtual (bottom) test scenarios under different environment conditions: clear (left), rain (middle) and night (right).} 
	\label{fig:scenarios}
\end{figure*}

Autonomous driving systems hold great potential to transform the transportation industry, and their development has been the focus of significant research efforts. However, ensuring the safe and reliable operation of autonomous vehicles under a wide range of environmental conditions, including adverse weather, is not possible yet due to the unreliable performance of the algorithms under adverse scenarios \cite{Zhang2021}. To aid in this effort, we have developed a  dataset that includes data of environment perception sensors (camera, radar and LiDAR) and  positioning systems (inertial measurement unit and GNSS). Tests were executed under adverse weather conditions (rain, night-time and snow) in two controlled environments: in a proving ground, and in laboratory. This means, the dataset contains sensor data acquired firstly in a real test track and sensor data acquired after the reproduction of the same test scenarios in a hardware-in-the-loop (HIL) test environment in laboratory. As an example, \Cref{fig:scenarios} illustrates a test scenario under three different weather conditions on the top row and its respective digital twin on the bottom row. 

The definition of the test scenarios was inspired by those catalogued by the European New Car Assessment Programme (EURONCAP) \cite{euroncap} and contain different types of objects including pedestrian, cyclist, truck, and car, each equipped with location sensors to provide accurate ground-truth data. In the laboratory, a digital twin of the real tests in the proving ground was developed for the reproduction of the tests in HIL. In this test environment, the same hardware and software stacks are used to generate the sensor data in laboratory with the help of environment simulation. For the stimulation of the perception sensors, different methods for injecting synthetic sensor data in the hardware of camera and radar sensors are implemented. 

\textbf{Related Works}.
The KITTI dataset \cite{Geiger2013} has played a significant role in the development of autonomous vehicles in the last decade by providing camera images alongside LiDAR and IMU data. Since then, multiple other datasets have been released. The Waymo Dataset \cite{waymo} provides a large amount of annotated camera and LiDAR data. The nuScenes dataset \cite{Caesar2019}  provides data from six cameras, one LiDAR, IMU, and five radars. This dataset contains 100 times more images than the KITTI dataset. Due to the cost and limitations of creating a dataset, the All-In-One-Drive \cite{Weng2020_AIODrive} is a dataset with synthetic data. It contains simulated scenes with collisions, high-speed driving, and violations of traffic norms. The Canadian Adverse Driving Conditions Dataset \cite{Pitropov2020}  is specifically designed to address adverse driving conditions, offering a diverse range of challenging scenarios. These scenarios encompass various weather conditions, road surfaces, and lighting conditions, ensuring a realistic representation of the complexities encountered during adverse driving situations. From heavy rain and fog to snow-covered roads and low-light environments. This dataset captures the intricacies and uncertainties present in adverse driving conditions with eight cameras, LiDAR and GPS/IMU.  
The proposed TWICE dataset provides both real sensor data and synthetic sensor data generated in HIL in clear and adverse weather conditions.

In this paper, we provide a detailed description of the dataset and its data collection process. 
Firstly, we present sensor setup used in the tests and their calibration methods. Following this, we explain the process of data collection, both on the real proving ground and in the laboratory. Then, followed by a discussion of the data processing, including data annotation and the data structure. We then present the recorded test scenarios and provide an overview of the dataset statistics. Finally, we provide an example of a python script on how to use the data.
\section{Data Collection}
\subsection{Sensors setup}
\begin{figure}[b!]
    \centering
    \includegraphics[width=\linewidth]{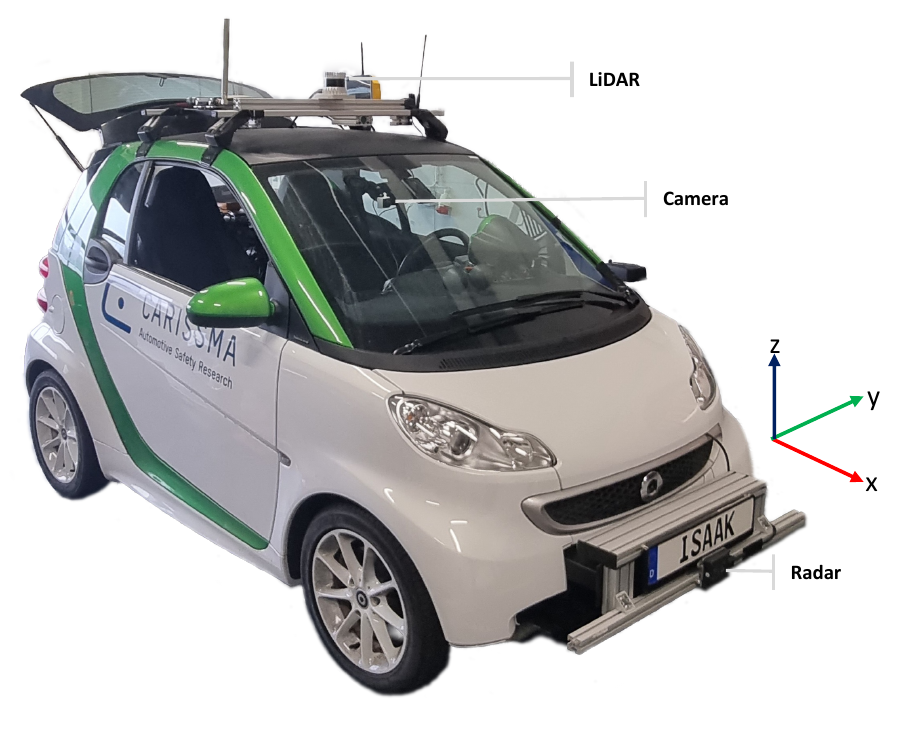}
    \caption{Ego-vehicle equipped with the sensors used for the test execution in the real proving ground.}
    \label{fig:sensors}
\end{figure}

The sensor setup, depicted in \Cref{fig:sensors}, was installed on a Smart ForTwo electric vehicle and includes:

\begin{itemize}
\item \textbf{Camera} Sekonix SF3325-100, 1920x1232 resolution (2.3 MP), \SI{30}{\hertz} capture frequency, and field of views of \SI{60}{\degree} (horizontal) and \SI{36,6}{\degree} (vertical);
\item \textbf{Radar} Continental ARS408-21, FMCW,  $\le$ \SI{250}{\m} range, \SI{13}{\hertz} capture frequency, \SI{77}{\giga \hertz} operating frequency band, ± \SI{0.1}{\kilo\meter\per\hour} velocity accuracy;
\item \textbf{LiDAR} Ouster OS1-128, \SI{10}{\hertz} capture frequency, resolution 1024.
\item \textbf{Inertial Measurement Unit} (IMU) Gensys ADMA-G PRO, 100Hz capture frequency, resolution \SI{0.01}{\m}/\SI{0.005}{\degree}; 
\item \textbf{GNSS} (Global Navigation Satellite System) SP80, resolution \SI{0.008}{\meter}, \SI{10}{\hertz} capture frequency;
\end{itemize}

The data is acquired with the Nvidia DRIVE PX 2, in which all sensors are connected and integrated using the Robotic Operating System (ROS) \cite{ROS.2018}. The radar was attached to the front bumper, the camera inside the car attached to the windshield, the LiDAR was mounted on the car's roof, and the IMU in the trunk alongside the ECU in the midpoint of the rear axle. 

To capture the precise location of the pedestrian, cyclist, and truck, we utilized GNSS SP80 sensor. When the object of interest was a car, we used an identical Smart ForTwo vehicle equipped with the same IMU. 

The radar data was acquired utilizing two distinct operational modes: the ``cluster'' mode, wherein each point represents a radar reflection within a point cloud, and the ``object list'' mode, where such reflections are internally to the radar ECU aggregated into bounding box representations of objects. In both cases, radar detections have velocity, position, and RCS (radar cross section).

The sensors which are installed in the ego vehicle have their position at the first moment unknown, therefore it is required that intrinsic and extrinsic calibrations are done.

\subsection{Sensors calibration}
\textbf{Extrinsic.} 
For the extrinsic calibration, methods vary by sensor.
First, the camera-to-vehicle calibration was performed following the method proposed in the Nvidia DriveWorks framework \cite{driveworks}.
Then, the radar-to-camera and LiDAR-to-camera calibration was performed with the help of a joint extrinsic calibration tool proposed in \cite{Domhof2019}.
This calibration technique requires a common reference target that can be detected by camera, radar and LiDAR sensors. 
Thereby the position of the camera, radar and LiDAR in relation to the vehicle can be jointly estimated.
For the IMU calibration, laser measurement tools were performed so that the position of the IMU is determined -- according to ISO 8855:2011 \cite{iso_axle}, which specifies the midpoint of the rear axle as reference.
\begin{figure}[t!]
    \centering
    \includegraphics[width=\linewidth]{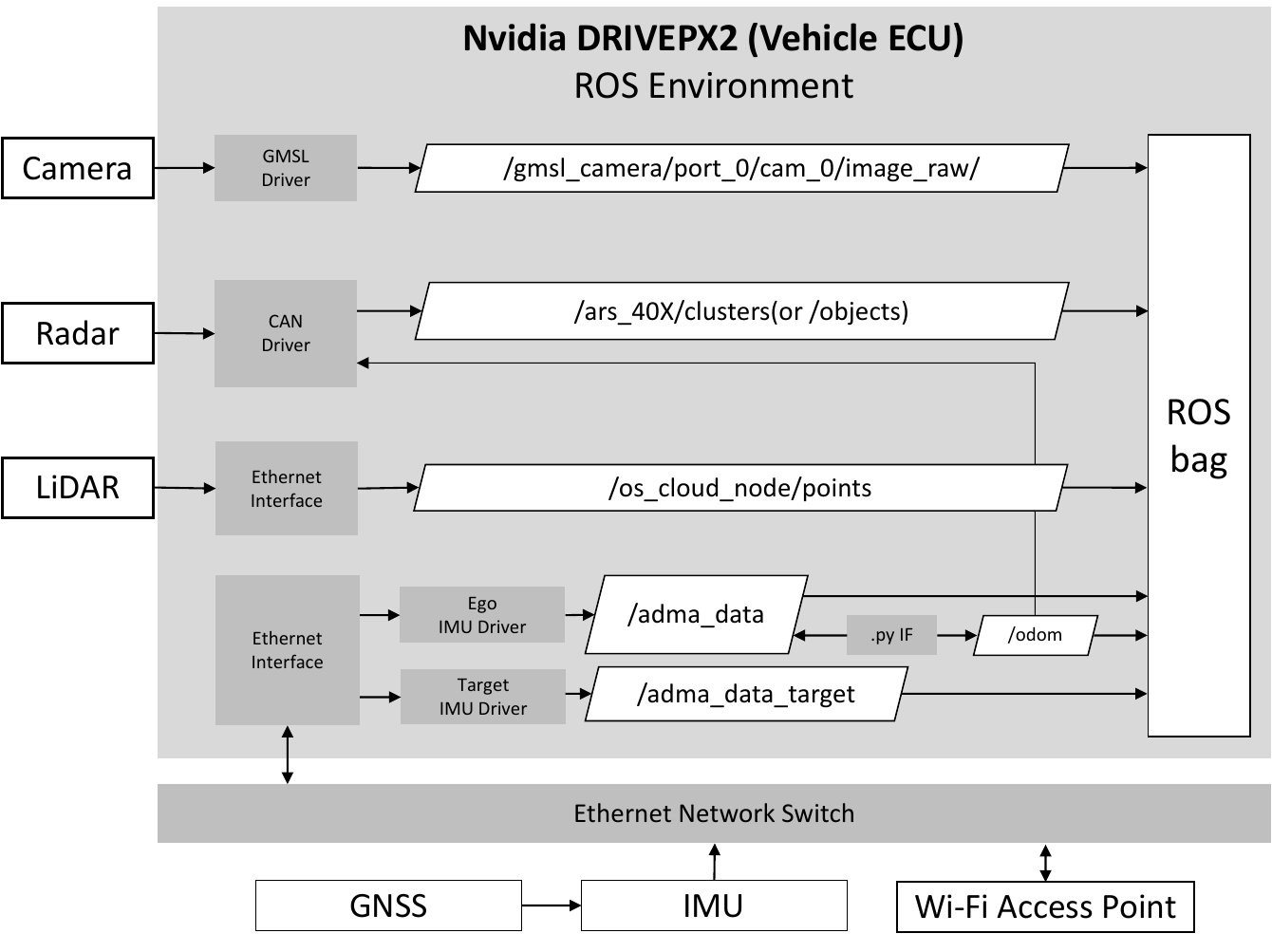}
    \caption{ROS environment for data acquisition.}
    \label{fig:ROS}
\end{figure}

\begin{figure}[b]
    \centering
    \includegraphics[width=\linewidth]{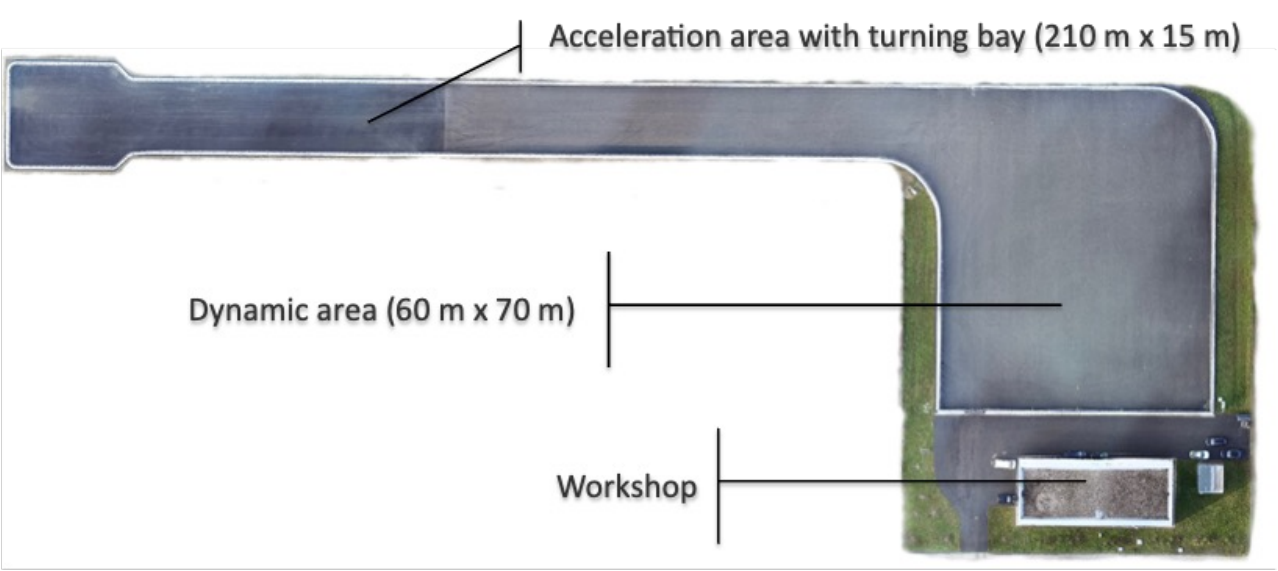}
    \caption{The CARISSMA outdoor proving ground.}
    \label{fig:outdoor}
\end{figure}

\subsection{Data recording}

\textbf{Real proving ground.}
For the purpose of sensor data collection, ROS was used as our main framework to simultaneously record data from various sensors during their operation. Each sensor creates a topic where its data was shared in real-time. This data was then stored in a ROSbag, as illustrated in \Cref{fig:ROS}, which depicts the logical architecture of our measurement system. Sensors attached to the ego-vehicle were directly connected to the ECU, while those used in  object positioning transmitted their data through an Ethernet network.

The CARISSMA Institute for Automated Driving (C-IAD) has an outdoor testing facility, illustrated in \Cref{fig:outdoor}, that was utilized as a proving ground for conducting our experimental scenarios. This testing facility is situated in Ingolstadt, Germany, and covers an area of 215m in length and 15m in width. In addition to this, there is a dynamic testing region measuring 60 meters by 70 meters.
\begin{figure}[t!]
    \centering
    \includegraphics[width=\linewidth]{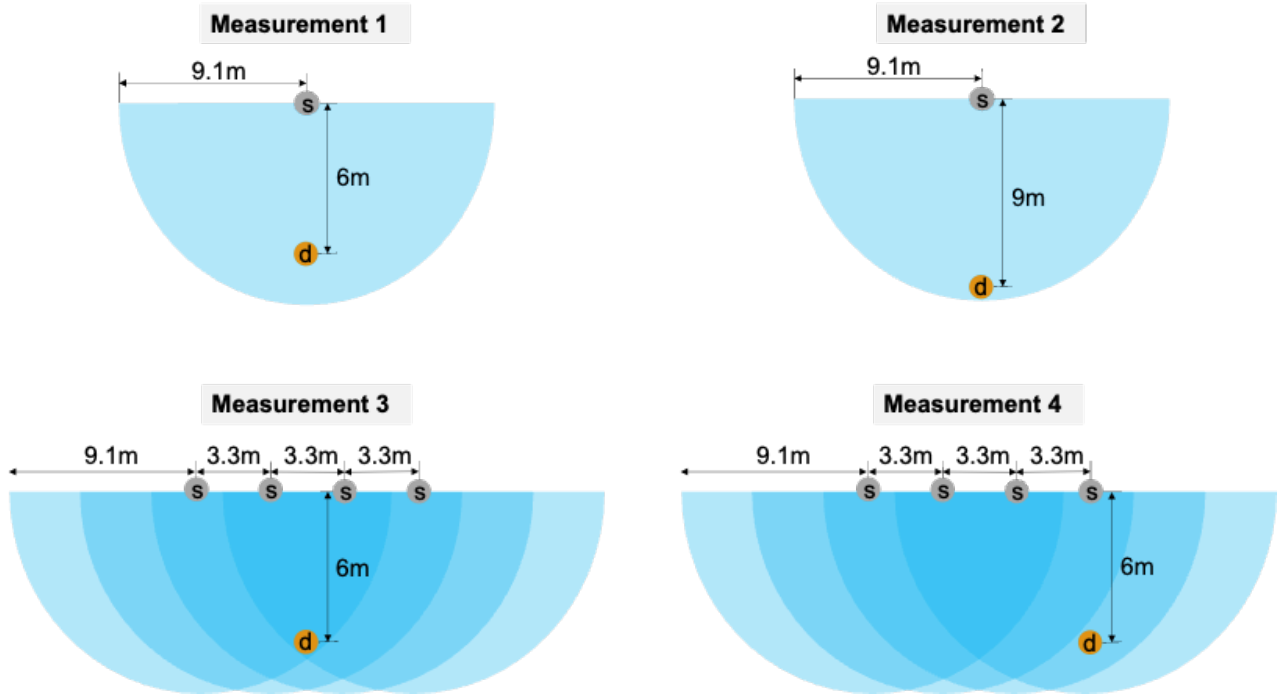}
    \caption{Disposal of the sprinkler system during the rain measurement. The grey dots are sprinkles and the orange dot is the measurement point.}
    \label{fig:sprinklers}
\end{figure}
\begin{figure}[b!]
    \centering
    \includegraphics[width=\linewidth]{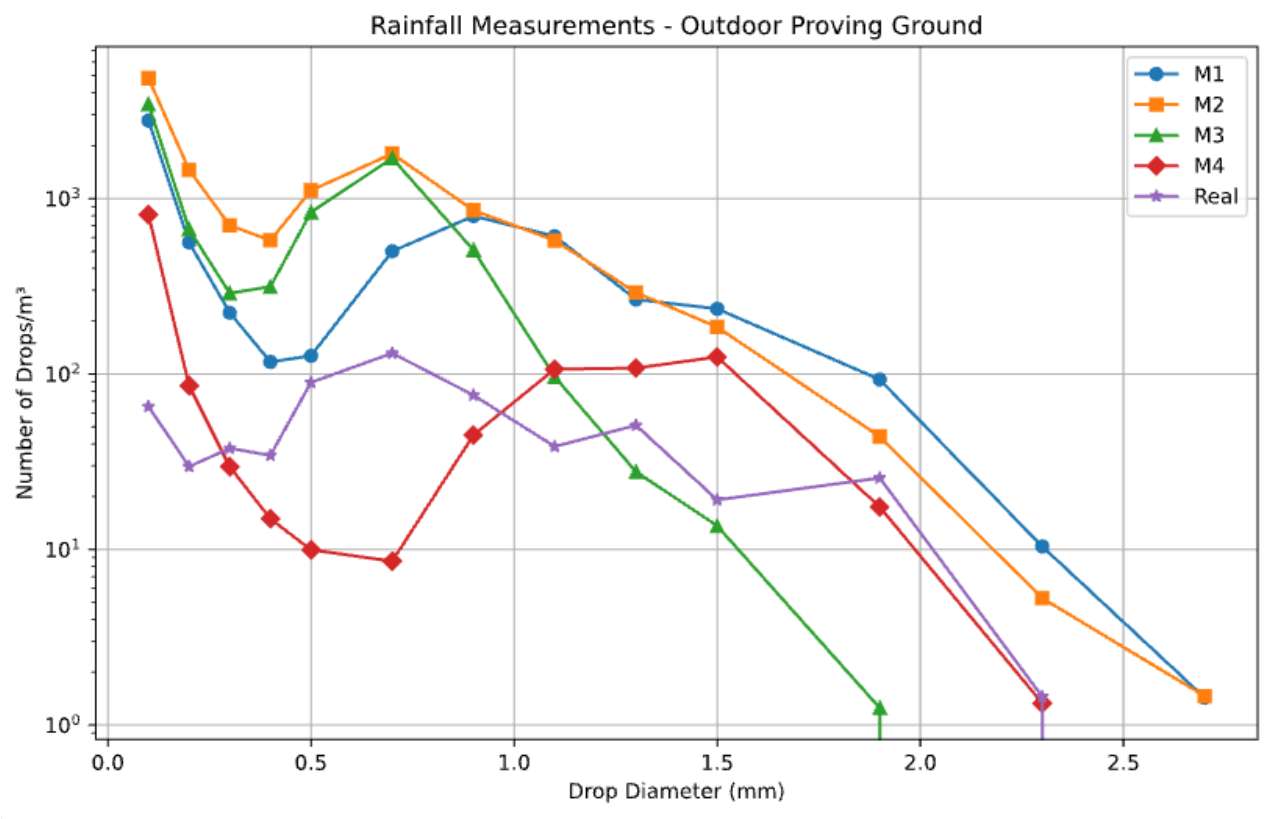}
    \caption{Raindrop size distribution (DSD) measured under different configurations, as well as a comparison with a DSD of real rainfall.}
    \label{fig:rain_graphic}
\end{figure}

\textbf{Reproducible Rain in the Proving Ground.}
In order to conduct experiments under rainy conditions, an artificial rain generation system was utilized, which allowed for controlled generation of rainfall. This system involved the use of a sprinkler system that was able to generate artificial raindrops. In the Figure \ref{fig:sprinklers} is shown the different configurations of the sprinklers system. 
Figure \ref{fig:rain_graphic} displays the raindrop size distribution (DSD) generated by this system measured under different configurations, as well as a comparison with a DSD of real rainfall.

\begin{figure}[t!]
    \centering
    \includegraphics[width=\linewidth]{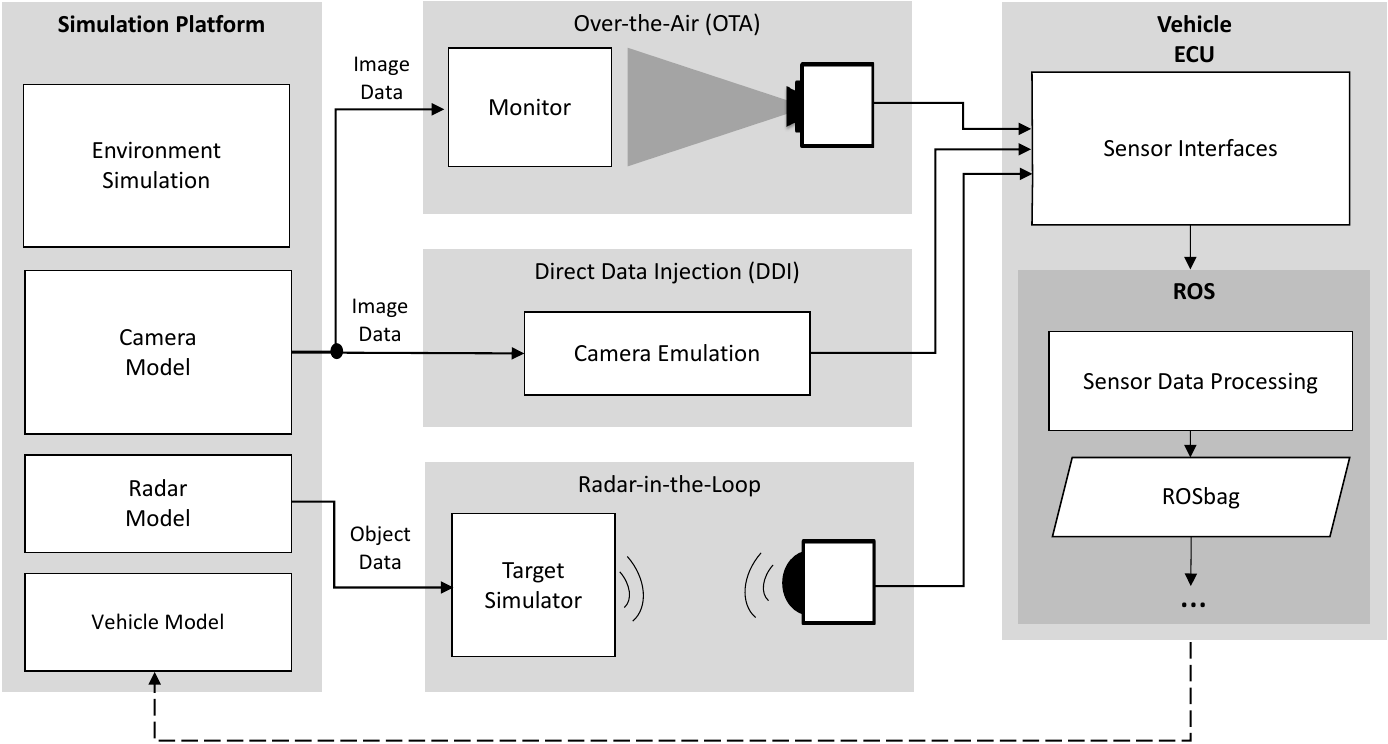}
    \caption{Data acquisition in the Laboratory: Stimulation of environment sensors (virtual tests, but integrated hardware of camera and radar).}
    \label{fig:hil}
\end{figure}

\textbf{Laboratory.}
In the laboratory, the HIL test environment -- illustrated in in \Cref{fig:hil} -- consists of the same hardware and software stack for the environment perception sensors (except the LiDAR and IMU), and includes the following additional equipment:

\begin{itemize}[leftmargin=*]
\item NI PXIe-8115 Controller;
\item NI PXIe-1085 Chassis with CAN modules (National Instruments);
\item Vehicle Radar Test System (VRTS) (Konrad Technologies);
\item Video Interface Box (IPG Automotive);
\item Camera-Box \cite{Reway_2018};
\item Environment Simulation Software CarMaker (IPG Automotive).
\end{itemize}

As it can be seen, the setup is a hardware-in-the-Loop test environment, in which real  sensors (radar and camera) are stimulated so that they can perceive the objects in the environment simulation software (CarMaker).

We employed two different methods for the stimulation of the camera sensor. The first type is defined as Direct Data Injection (DDI) and it uses the Video Interface Box -- a middleware to transfer synthetic image data from the simulation to the ECU over cable. The second type is defined as Over-the-Air (OTA), in which a camera is employed to capture a high-resolution display during the execution of a simulation. \cite{Reway2018, Reway_2022}.
In the case of the DDI, there was no distortion present, whereas the OTA setup exhibited significant distortion. Thefore, the intrinsic calibration has to be performed once again for the OTA setup.

For the radar stimulation, an anechoic chamber was built, in which the radar was mounted in one side, and the VRTS antennas on the opposite side. The VRTS receives the object data from the simulation, and, by manipulating the radar waves, it is able to generate a reflection to emulate a target for the radar \cite{Kizilay.76}. 
The mathematical definition of radar wave manipulation for radar target simulation is detailed in \cite{Gruber_2017}.

\section{Dataset data formating}

\subsection{Data pre-processing}
During the process of collecting ground truth data for the objects of interest, the GNSS SP80 sensor has a signal loss in some scenarios. To mitigate this issue, we utilized data interpolation techniques. Each of the interpolated data is accompanied by a corresponding \texttt{.txt} file that specifies the exact moment in the test scenario in which the interpolation was applied.

For scenarios where the object of interest was a car, the IMU was employed. Although in a few cases there was a lack of data, this was minimal, and therefore, data interpolation was not necessary. The precise moments in the test scenario in which data was lacking are also specified in the corresponding \texttt{.txt} file.

\begin{figure}[b!]
    \centering
    \includegraphics[width=\linewidth]{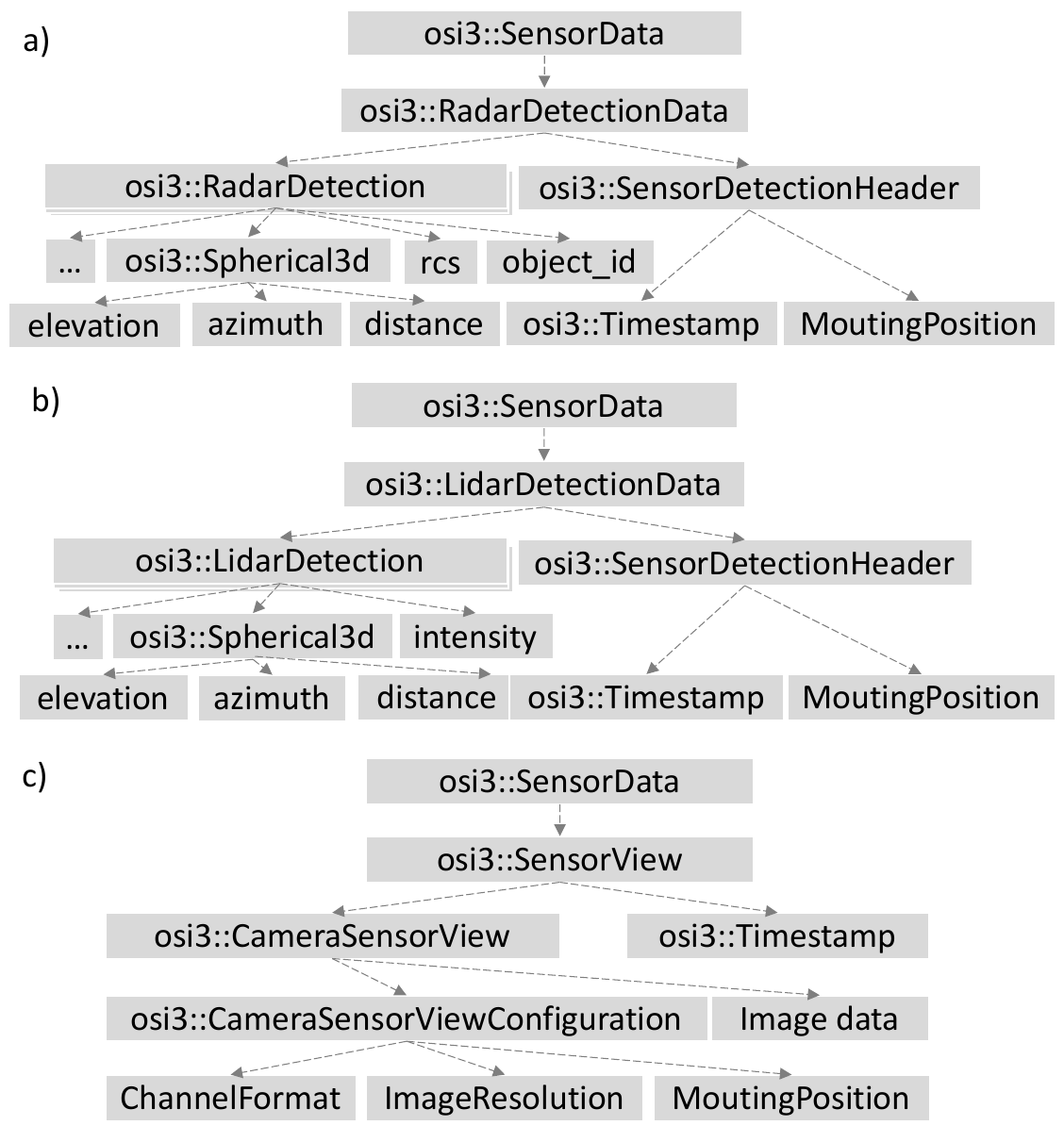}
    \caption{Structure of OSI classes. How the sensor data is stored: Radar (a), LiDAR (b), Camera (c)}
    \label{fig:osi}
\end{figure}

\subsection{Dataset data format}

The Open Simulation Interface (OSI) \cite{OSIreport}, maintained by the Association for Standardization of Automation and Measuring Systems (ASAM), is a standardized interface designed for the intercommunication of various components within a simulation environment. We utilized OSI to structure our sensor data, leveraging its efficacy in organizing and preserving vast data volumes through a language-agnostic serialization method: Google Protocol Buffers (protobuf). This approach aligns with the emerging ISO 23150 standard \cite{iso}, promoting standardized communication interfaces for real sensors. Such alignment guarantees the modularity, integrability, and interchangeability of each component. Using OSI, we can archive both the sensor data and the measured object position (ground-truth) for a given scenario.

Each sensor has an individual \texttt{.osi} file assigned to it. In the case of the radar, IMU, and ground-truth of objects, the \texttt{SensorDataSeries} is utilized, which generates a list of detections contained within the protobuf message. However, for the Camera and LiDAR sensors, the protobuf has a default size limit of 64MB. As a result, each detection is stored in its individual protobuf message and subsequently parsed into a single .osi file using the Python's \texttt{struct} package.

The Figure \ref{fig:osi} exemplifies how the data is stored in the OSI format. This structure repeats for each frame resulting in a protobuf object to be stored in a single \texttt{.osi} file.

The data is stored within a directory structure that divides the tests by scenario, whether the data is real or synthetic, weather condition and radar operation mode.

\begin{figure}[h!]

\begin{lstlisting}[language=json,firstnumber=1]
{
  "openlabel": {
    "metadata": {...},
    "frames": [
      {
        "frame_properties": {
          "timestamp": 0.01425004,
          "Streams": {
            "Camera1": {
              "stream_properties": {...},
                "intrinsics_pinhole": {
                  "camera_matrix_3x4": [...],
                    (...)
                    }
              }
            },
            "IMU_ego": {
              "stream_properties": {
                "sync": {
                  "frame_stream": 1,
                  "timestamp": 0.009988784
                }
              }
            },
            "IMU_obj": {...},
            "Radar": {...}
            }
          }
        },
        "objects": [
          {
            "name": "Car_1",
            "type": "Car",
            "object_data": {
              "cuboid": {
                "name": "cuboid",
                "val": [... ]
              }
            }
          }
        ]
      }
\end{lstlisting}
    \caption{OpenLABEL structure example.}
    \label{fig:open_label}
\end{figure}
\subsection{Data annotation}

The labeling of camera data was accomplished by projecting the ground-truth of the object in the camera frames. Based on ego position, object position, and camera intrinsic/extrinsic parameters, the 3D cuboid can be generated and then projected onto the desired image frame. To validate the accuracy of the ground truth projection, YOLOv4 \cite{YOLOv4} was used for supervision. The resulting bounding box was used to project the cuboid and compute the Intersection-over-Union (IoU). Any necessary adjustments in the ground-truth position were made based on the IoU analysis, thereby ensuring precise labeling of the object.

The labeling data was stored utilizing the ASAM OpenLABEL \cite{openlabel}, which defines the annotation format and labeling methods for objects and scenarios. This framework, developed by ASAM, enabled us to store both the cuboid information and synchronization between the sensors in a \texttt{}{.json} file, as illustrated in Figure \ref{fig:open_label}.
The timestamp of all sensors in our study is based on the ROS timestamp. Therefore, synchronization of the sensors involved identifying the closest timestamp of each sensor and creating an index list. Both the camera and radar sensors each had their own OpenLABEL file.

\section{Scenarios}

In this section, the scenarios are discussed. The use of test scenarios inspired by the EURONCAP test protocols enables valid and useful comparisons between tests in different facilities and weather conditions. 

\begin{figure}[b!]
    \centering
    \includegraphics[width=\linewidth]{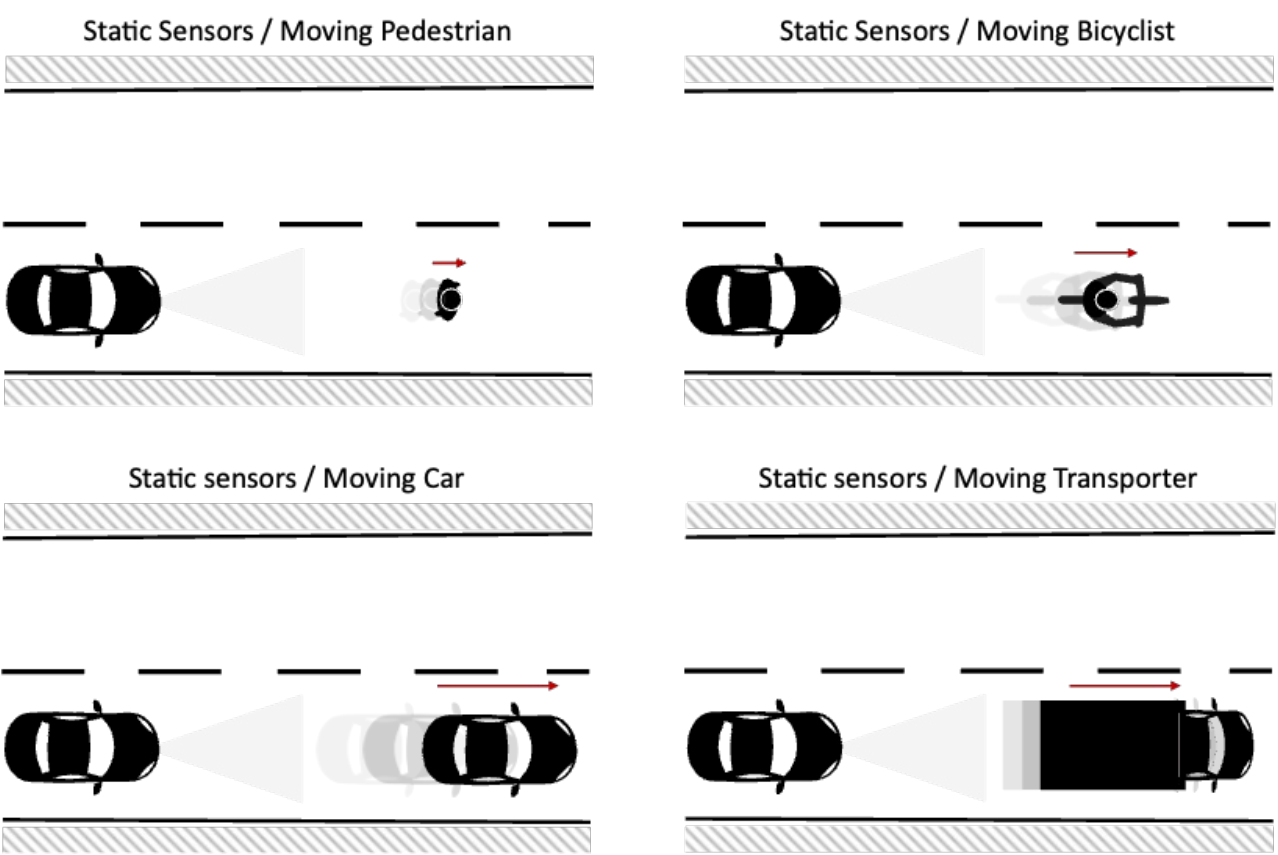}
    \caption{With the ego-vehicle static 4 types of objects moving away.}
    \label{fig:static_ego}
\end{figure}

\subsection{EURONCAP Scenarios}
In this work we recorded eight different test scenarios, three of which are inspired in the EURONCAP test protocols:
\begin{itemize}[leftmargin=*]
\item \textbf{Car-to-Bicyclist Longitudinal Adult 50\% (CBLA-50)}: a collision in which a vehicle travels forwards towards a bicyclist cycling in the same direction in front of the vehicle where the vehicle would strike the cyclist at 50\% of the vehicle’s width when no braking action is applied.\cite{euroncap}

\item \textbf{Car-to-Car Rear Braking (CCRb)}: a collision in which a vehicle travels forwards towards another vehicle that is travelling at constant speed and then decelerates, and the frontal structure of the vehicle strikes the rear structure of the other.\cite{euroncap2}

\item \textbf{Car-to-Car Rear Stationary (CCRs)}: a collision in which a vehicle travels forwards towards another stationary vehicle and the frontal structure of the vehicle strikes the rear structure of the other.\cite{euroncap2}
\end{itemize}
In the CBLA test scenario an EURONCAP certified test target was utilized, representing a bicyclist. Ideally, the ground sensor would be in the target itself, however the test target was pulled by a car equipped with an IMU sensor to obtain the ground-truth. In the case of the CCRs, was used an EURONCAP certified test target representing a car, with the position fixed during all tests. All the test scenarios that involved a car as the object of interest utilized a vehicle identical to the ego, including the CCRb.

\subsection{Other scenarios}

With the ego-vehicle static was recorded four types of objects:  car, cyclist, truck and pedestrian. As it can be seen in Figure \ref{fig:static_ego}. The recordings were made with the objects initially positioned in front of the ego-vehicle and then moving away. For the bicyclist, pedestrian and truck the GNSS SP80 sensor was utilized in order to get the ground-truth position. The last of the eight different test scenarios involves the ego-vehicle traveling forward towards a static truck that is positioned perpendicular to the ego-vehicle.

\begin{table}[t]
\caption{Scenarios distribution.}
\label{tab:scenarios}
\begin{adjustbox}
{width=\columnwidth,center}
    \centering
    \large

\begin{tabular}{c|cc|cc|cc|cc}
                    & \multicolumn{2}{c}{daytime} & \multicolumn{2}{c}{night} & \multicolumn{2}{c}{rain} & \multicolumn{2}{c}{snow} \\
Type                & real       & synthetic      & real      & synthetic     & real     & synthetic     & real     & synthetic     \\ \hline
CBLA                & 7          & 14             & 6         & 12            & 6        & 12            & -        & -             \\
CCRb                & 6          & 6              & 6         & 6             & 6        & 6             & -        & -             \\
CCRs                & 5          & 5              & 6         & 6             & 6        & 6             & -        & -             \\
Car                 & 6          & 6              & 6         & 6             & 6        & 6             & 4        & 4             \\
Bike                & 6          & 6              & 5         & 5             & 6        & 6             & 4        & 4             \\
Pedestrian          & 6          & 6              & 6         & 6             & 6        & 6             & 1        & 1             \\
Truck               & 4          & 4              & 3         & 3             & -        & -             & 4        & 4             \\
Truck perpendicular & -          & -              & 4         & 4             & -        & -             & 4        & 4     
\end{tabular}
\end{adjustbox}

\end{table}

\subsection{Statistics}

The dataset contains 221.495 camera frames, which 99.516 are collected in the real proving ground and 121.979 generated in the laboratory. This frames make part of 289 test-runs, that totals more than 2 hours of recording. 

\begin{figure}[b]
    \centering
    \includegraphics[width=\linewidth]{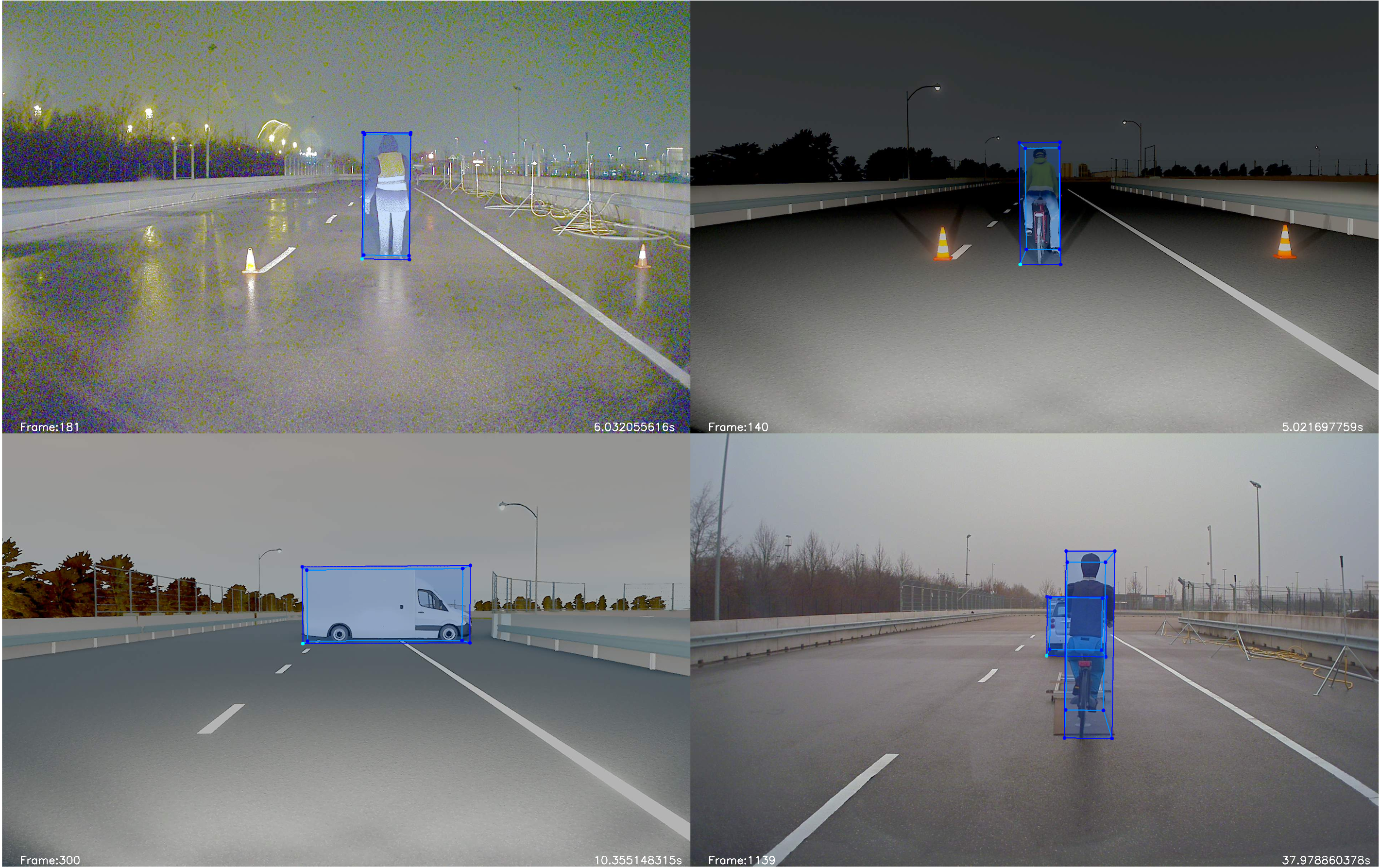}
    \caption{Cuboids projected in four types of objects. }
    \label{fig:cuboids}
\end{figure}

\Cref{tab:scenarios} displays the distribution of test runs based on scenarios and weather conditions. Of these, 135 were recorded on the real proving ground, while 154 were conducted in the laboratory. The increased number of synthetic test runs in the laboratory is attributed to the CBLA tests. These tests were executed both with and without the car towing the bicyclist.

\section{How to use the data}

Sensor information is stored in OSI files to be easily accessible. 
Extensive documentation on the structure
and format is available on the OSI website \cite{OSIsite}. We also provide a python notebook that allows to open and visualize all the data. It is possible to project the cuboids, which are stored in the OpenLABEL files, onto image frames and generate a video file with the ground-truth (Figure \ref{fig:cuboids}).

\begin{figure}[b!]
    \centering
    \includegraphics[width=.9\linewidth]{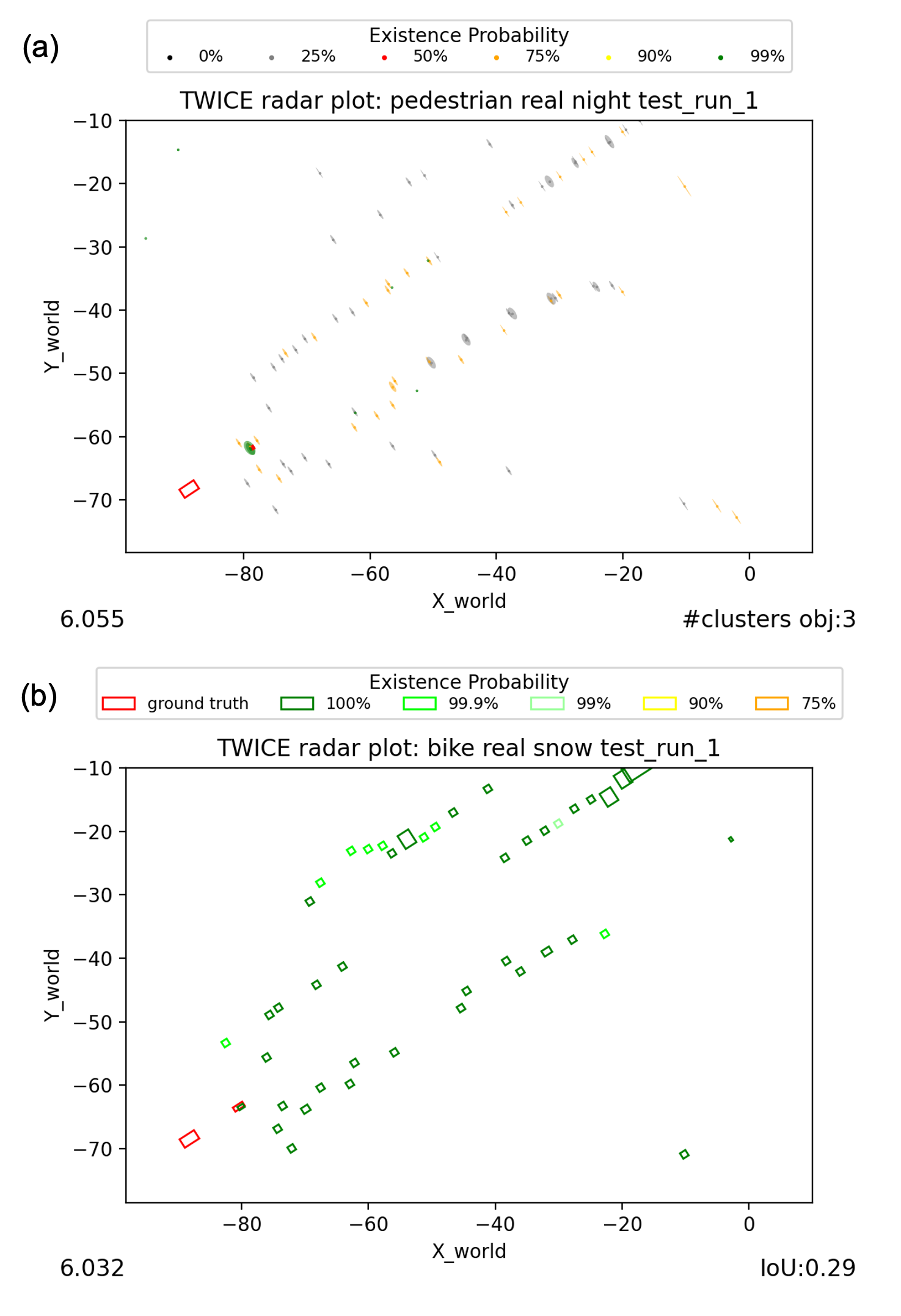}
    \caption{Radar plot alongside with the position of the ego and objects. Operation mode: (a) Cluster (b) Object List}
    \label{fig:radar_camera}
\end{figure}

The radar data can be visualized together with the position of the ego and the object of interest, as shown in Figure \ref{fig:radar_camera} (a) for the cluster data and (b) for object list data.

\subsection{Folder structure}

\begin{figure}[t!]
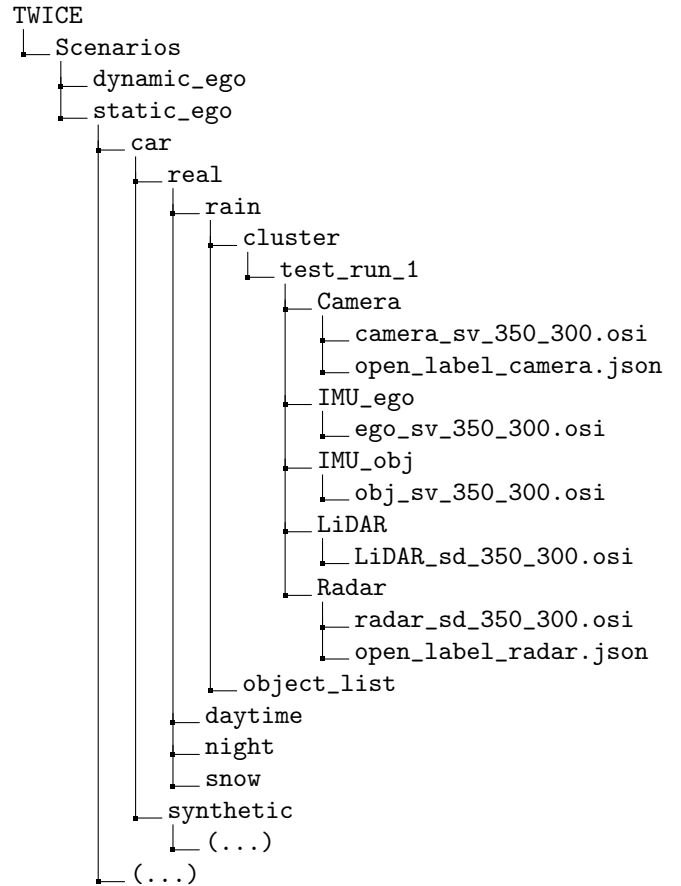


\dirtree{%
 .1 TWICE.
 .2 Scenarios.
 .3 dynamic\_ego.
 .3 static\_ego.
 .4 car.
 .5 real.
 .6 rain.
 .7 cluster.
 .8 test\_run\_1.
 .9 Camera.
 .10 camera\_sv\_350\_300.osi.
 .10 open\_label\_camera.json.
 .9 IMU\_ego.
 .10 ego\_sv\_350\_300.osi.
 .9 IMU\_obj.
 .10 obj\_sv\_350\_300.osi.
 .9 LiDAR.
 .10 LiDAR\_sd\_350\_300.osi.
 .9 Radar.
 .10 radar\_sd\_350\_300.osi.
 .10 open\_label\_radar.json.
 .7 object\_list.
 .6 daytime.
 .6 night.
 .6 snow.
 .5 synthetic.
 .6 (...).
 .4 (...).
}
\caption{Folder structure of the TWICE dataset.}
\label{fig:folder_struct}
\end{figure}

The folder structure is illustrated in \Cref{fig:folder_struct}, which is crucial for ensuring systematic data access, comprehension, and efficient subsequent data processing for any interested party.

\section{Conclusion}
We presented a unique dataset that contains data from camera, radar, LiDAR and IMU. Recorded in a controlled environment under adverse weather conditions. For each test scenario on the real proving ground there is a digital twin with two types of camera (DDI and OTA), IMU data, ground-truth of objects and radar data recorded in a HiL environment using the same hardware as the real test.

As simulations are increasingly employed for testing of automated driving algorithms \cite{Böde2018, Riener.2016}, the proposed TWICE dataset offers a potential avenue to explore the simulation-to-reality gap, facilitating validation of simulations and thereby saving considerable time and resources in the evaluation of these systems.

The dataset contains 8 different test scenarios, three of which are inspired by EURONCAP test scenarios, and were recorded under four types of weather conditions: daytime, nighttime, rain, and snow. It is more than 280 GB of data (80 GB compressed).

\section*{Acknowledgments}
{\small
This research was supported by BayWISS-Verbundkolleg
”Mobility and Transport”, the AWARE Center and its project
enGlobe – Engineers go global, funded by the German Academic Exchange Service (DAAD) and Federal Ministry of
Science and Research (BMBF)
and was financed in part by the Coordenação de
Aperfeiçoamento de Pessoal de Nível Superior - Brasil (CAPES) - Finance Code 001
}
\renewcommand{\bibfont}{\footnotesize}
\printbibliography 


\end{document}